# Bias in Large Language Models Across Clinical Applications: A Systematic Review

Thanathip Suenghataiphorn[1], Narisara Tribuddharat[2], Pojsakorn Danpanichkul[3], Narathorn Kulthamrongsri[4]

[1]Department of Internal Medicine, Griffin Hospital, Derby, CT, United States, [2]St. Elizabeth Medical, Boston, MA, United States, [3]Department of Internal Medicine, Texas Tech University Health Science Center, Lubbock, TX, United States and [4]University of Hawaii, Honolulu, HI, United States

Corresponding Author: Thanathip Suenghataiphorn, Department of Internal Medicine, Griffin Hospital, Derby, CT; 130 Division St, Derby, CT, United States 06418; Thanathip.sue@gmail.com; 443-484-8464

**Abstract Word Count**: 244 words

**Word count:** 2,786 words

**Tables:** 2; **Figures:** 2

**Conflict of Interest**: We declared no conflict of interest

**Fundings:** No funding was sought for this study

**Acknowledgments/Disclosure**: During the preparation of this work the authors used Gemini in order to summarize each original article. After using this tool/service, the authors reviewed and edited the content as needed and takes full responsibility for the content of the publication.

Ethics Statement: This study does not involve human participants and this study does not involve animal subjects

**Author contributions**

Conceptualization – TSu, NT

Data curation – TSu, NK

Formal analysis – TSu, PD

Investigation (Search) – TSu, PD, NT

Methodology – Tsu; Validation – Tsu

Writing, original draft – TSu, NK

Finalized the manuscript – TSu, PD

**All authors have read and approved the final version of the manuscript for submission.**

## ABSTRACT

**Background:** Large language models (LLMs) are rapidly being integrated into healthcare, promising to enhance various clinical tasks. However, concerns exist regarding their potential for bias, which could compromise patient care and exacerbate health inequities. This systematic review investigates the prevalence, sources, manifestations, and clinical implications of bias in LLMs.

**Methods:** We conducted a systematic search of PubMed, OVID, and EMBASE from database inception through 2025, for studies evaluating bias in LLMs applied to clinical tasks. We extracted data on LLM type, bias source (data-related or model-related), bias manifestation (allocative harm, representational harm, or performance disparities), affected attributes, clinical task, evaluation methods, and outcomes. Risk of bias was assessed using a modified ROBINS-I tool.

**Results**: Thirty-eight studies met inclusion criteria, revealing pervasive bias across various LLMs and clinical applications. Both data-related bias (from biased training data) and model-related bias (from model training) were significant contributors. Biases manifested as: allocative harm (e.g., differential treatment recommendations); representational harm (e.g., stereotypical associations, biased image generation); and performance disparities (e.g., variable output quality). These biases affected multiple attributes, most frequently race/ethnicity and gender, but also age, disability, and language.

**Conclusions**: Bias in clinical LLMs is a pervasive and systemic issue, with a potential to lead to misdiagnosis and inappropriate treatment, particularly for marginalized patient populations. Rigorous evaluation of the model is crucial. Furthermore, the development and implementation of effective mitigation strategies, coupled with continuous monitoring in real-world clinical settings, are essential to ensure the safe, equitable, and trustworthy deployment of LLMs in healthcare.

## STUDY HIGHLIGHTS

WHAT IS KNOWN

- Large Language Models (LLMs) are increasingly being integrated into healthcare, offering potential benefits across a spectrum of applications, from streamlining administrative workflows to assisting in complex clinical decision-making.
- Bias in artificial intelligence, particularly in machine learning models, is a well-documented issue, with prior research demonstrating its presence in various domains, including some initial investigations in non-clinical LLM applications. Previous clinical studies are limited in scope and sample size.

WHAT IS NEW HERE

- This systematic review provides the first comprehensive evidence that bias, particularly stemming from data-related and model-related sources, is a pervasive and systemic issue in LLMs applied to a broad range of clinically relevant tasks. This bias has the potential to amplify them, leading to misdiagnosis, delayed treatment, undertreatment of pain, and ultimately, poorer health outcomes for marginalized patient populations.
- The identified biases manifest as allocative harm, representational harm, and performance disparities, affecting critical clinical processes such as diagnostic reasoning, treatment recommendations, and the generation of clinical documentation. These biases are observed across a range of attributes including, but not limited to race, ethnicity, gender, age, and socioeconomic status.
- Mitigation strategies, while nascent, must address bias at multiple levels, including data curation, model development, and clinical deployment, with continuous monitoring and evaluation being essential. Continuous, rigorous evaluation of LLMs in real-world clinical settings is essential to ensure their safe, equitable, and trustworthy deployment. Future research needs to go beyond identifying bias and develop, test and validate tools for mitigation and reduction of bias across the different categories of bias.

## INTRODUCTION

Large language models (LLMs) are rapidly transforming the landscape of artificial intelligence, demonstrating remarkable capabilities in natural language processing, generation, and understanding. These powerful models, trained on vast corpora of text and code, are increasingly being deployed across diverse sectors, with healthcare emerging as a particularly promising, yet challenging, frontier[1]. From automating administrative tasks and summarizing patient records to assisting with clinical decision-making and generating personalized treatment plans[2], LLMs hold the potential to revolutionize healthcare delivery, improve efficiency, and enhance patient outcomes.

However, LLMs usage also raises critical concerns, particularly regarding the potential for bias[3]. LLMs learn from the data they are trained on, and this data often reflects existing societal biases, including those related to race, ethnicity, gender, socioeconomic status, age, sexual orientation, and disability[4]. When these biases are encoded within LLMs, they can be amplified and perpetuated, leading to potentially harmful consequences, especially in the sensitive and high-stakes domain of healthcare[5]. For instance, a biased LLM might misdiagnose patients from underrepresented groups, recommend inappropriate treatments, or exacerbate existing health disparities.

This systematic review addresses this critical gap by comprehensively examining the existing literature on bias in LLMs, with a specific focus on clinical applications. We aim to synthesize the evidence on: (1) the types and sources of bias identified in clinically-relevant LLMs; (2) the methods used to detect and measure these biases; (3) the potential clinical consequences of biased LLM outputs; and (4) the proposed strategies for mitigating bias in LLMs for healthcare. By providing a systematic overview of the current state of knowledge, this review aims to inform the responsible development, deployment, and regulation of LLMs in healthcare, ensuring that these powerful technologies are used to promote equity, fairness, and improved patient care for all.

## METHODS

### *Search strategy & Eligibility*

Two investigators (TS and NT) independently conducted searches in PUBMED, OVID, and EMBASE databases from inception through 27th January 2025 using the search strategy as specified in **Supplementary Table 1**. The investigators (TS and NT) independently assessed the eligibility of the retrieved records. Any conflicts were resolved through further discussions involving a third investigator (PD). The protocol was designed based on the Preferred Reporting Items for Systematic Review and Meta-Analysis Protocols (PRISMA) checklist as seen in **Supplementary Table/Material 2**[6]. The protocol was preregistered (PROSPERO: 649773). Studies were included if they satisfied the following criteria:(1) any type of studies that investigated the potential bias role of LLMs in the medical, medicine or physician related (eg. studies that investigated bias generated from the LLM model, that was analyzed by comparing between ground truth and LLM result or between LLM models, are acceptable) (2) studies published in English, and (3) studies in full-text format. The exclusion criteria were: (1) studies that did not report on the direct application or potential role of LLMs in clinical medicine/physician (e.g., studies focusing solely on technical development without clinical application); (2) case reports, review papers; (3) studies involving non-human subjects (e.g., animal or basic research); (4) studies that focus solely on accuracy of the response; and (5) studies with insufficient data regarding the evaluation of bias in LLMs. Only publications conducted on human participants were included. The bibliographies of relevant papers were also reviewed to identify additional studies. No restrictions were applied regarding age, sex, or country of origin.

*Data extraction*

To ensure consistency and rigor, we implemented a standardized data collection protocol across all included studies. This protocol facilitated the systematic extraction of key variables pertinent to our research objectives. Extracted data included demographic information (first author's surname, country of study, year of publication), the primary AI/LLM model used, the study's topic and aims, evaluation criteria, the population or data source, the method of applying the LLM, the ground truth used for comparison, any other LLM models used in the study, and the main outcomes reported.

*Data synthesis*

Due to the heterogeneity in the specified aims and methodologies of the included studies, a narrative synthesis approach was employed. This involved a qualitative synthesis of the findings, categorized by the types of bias, impact, implications, and future considerations.

Due to the limited information in the types of bias exist from the large language model, we utilized a three-view of bias of evaluation, namely source of bias, manifestation of bias and attributes of bias. First, we categorize the source of bias studied into three types: Data-related bias, Model-related Bias and Deployment Bias. These three bias corresponding to the three key stages of constructing a large language model[7]: data collection and preparation, model training and testing, and post-deployment of the model. Second, we examined the manifestation or the harm of the bias. We assigned three types of harms as follows: allocative harm[8], representational harm[9] and performance disparities. Allocative harm occurs when the model systematically allocates resources or opportunities unfairly across different groups. Representational harm is assigned when the model reinforces negative stereotypes or diminishes the status of certain groups. We then allocated all other kinds of manifestation into performance disparities, as models can exhibit differences, but does not fall into the above categories. Lastly, the attributes of bias are documented, in which we focus on *which groups* are affected by the bias, such as race and ethnicity, gender or socioeconomic status.

*Methodological Quality Assessment*

The quality of the included studies was assessed using the ROBINS-I tool[10], which is designed for evaluating the risk of bias in non-randomized studies. This tool examines seven domains where bias can occur: confounding, participant selection, intervention classification, deviations from intended interventions, missing data, outcome measurement, and selective reporting of results. Each domain is rated as having a "low," "moderate," "serious," or "critical" risk of bias. The overall risk of bias for each study was determined by considering the highest risk rating across all domains. For instance, if a study had a "serious" risk in one domain and "moderate" or "low" in others, the overall risk was considered "serious." Two reviewers (TS and NT) independently assessed each study, and any disagreements were resolved through discussion or with the assistance of a third reviewer (PD).

## RESULTS

Our search strategy identified 1065 records. After removing 286 duplicates, we reviewed those studies by title and abstract, excluding 735 studies that did not meet the eligibility criteria related to study design or participants. Subsequently, we thoroughly reviewed 45 articles and excluded 7 for reporting different topic/domain. Ultimately, 38 studies met the eligibility criteria for our systematic review[11-48]. **Figure 1** illustrates our search methodology and selection process, and **Table 1** details each recorded selected papers. **Table 2** shows the methodological quality of each study as assess by ROBIN-I tool. **Figure 2** shows the heatmap of the amount of research studies, with source of bias and manifestation of bias.

*Source of Bias*

Data-related bias, originating from the datasets used to train the LLMs, emerged as a significant and recurring issue. This type of bias arises when the training data does not accurately reflect the real-world diversity of patient populations, clinical presentations, or relevant contextual factors. Several studies highlighted the detrimental effects of skewed datasets. For example, imbalances in the representation of skin tones in dermatological image datasets[48]. Studies analyzing the readability and content of LLM-generated text found that outputs varied significantly based on patient race[12] and socioeconomic status[29]. Furthermore, the utilization of historical data that reflects past societal biases, even if technically "accurate," can lead LLMs to reproduce and amplify those biases in their outputs[19,31,34,46]. This was evident in studies where models associated specific diagnoses with particular demographic groups, potentially influencing clinical decision-making and reinforcing harmful stereotypes[14,47]. None of the included studies directly assessed deployment bias, indicating a significant gap in the current literature.

Beyond the data itself, the design, training, and evaluation of LLMs introduce another significant source of bias: model-related bias. A striking example of model-related bias is the consistent tendency of image generation models to depict healthcare professionals, such as physicians and surgeons, as predominantly White and male[13,20,21,26,27]. This could be likely stemming from a combination of biased training data and the model's inherent tendencies. Furthermore, even when presented with identical inputs, different LLMs can exhibit varying degrees of bias, highlighting the influence of model-specific design choices. For instance, one study[45] demonstrated that Gemini was more likely than GPT-4 to recommend strong opioids, showcasing how algorithmic differences can lead to harm.

*Manifestation of Bias*

The biases stemming from both data and model sources are consistently observed across the reviewed studies. *Allocative Harm* was frequently seen in studies examining diagnostic pathways and treatment recommendations, where model outputs varied based on patient demographics[14,31,34,46] *Representational Harm* manifested as the reinforcement of harmful stereotypes or the under-representation of certain groups[13,18,24,25,33,35,43]. Finally, *Performance Disparities* were identified in some studies[15,42] . We noted that some studies reporting no significant bias[32,37,41,44].

*Attributes of Bias*

Race/ethnicity and gender were the most frequently investigated attributes, with numerous studies documenting biases related to these characteristics[11,12,18,19,21,24,25,31,33,35,36,39,40,46,48]. However, the scope of the reviewed studies extended to other crucial attributes, including age[14,18,21,27], socioeconomic status[29,34], disability status[43], language[42], and specific health conditions (smoking status)[15].

*Methodological Quality and Risk of Bias*

The overall methodological quality of the included studies was variable. Based on the ROBINS-I assessment, the majority of studies had a low risk of bias across most domains (**Table 2**). However, several studies exhibited a moderate risk of bias in the domains of outcome measurement and selective reporting of results. This was often due to the subjective nature of evaluating LLM-generated responses and the potential for selective reporting of positive findings.

## DISCUSSION

This systematic review, encompassing 38 studies of large language models (LLMs) applied to a range of clinical tasks, provides the first comprehensive overview of the complexity nature of bias within clinical application. The findings of this systematic review unequivocally demonstrate that bias in clinical LLMs is not a theoretical concern but a pervasive and systemic problem with the potential to significantly compromise patient care and exacerbate health inequities.

*Bias in Large Language Model*

The earliest studies of generative AI in Bias focuses on retrieving information, from typical large language models[20], to more specific fine-tuned models[44]. Data-related bias, the first major source, stems from the vast datasets that form the foundation of LLM training[49]. These datasets, often scraped from the internet or compiled from existing medical records[50], are rarely perfectly representative of the real-world diversity of patient populations and clinical scenarios[4], as seen in racial, disability and some individuals. Historical bias may explain this phenomenon. Medical data, by its very nature, reflects the history of medical practice, which has often been marked by disparities and inequities[51]. Past biases in diagnosis, treatment, and access to care are embedded within historical medical records, and if these records are used to train LLMs without careful consideration, the models will inevitably learn and perpetuate these historical injustices.

Model-related bias, the second primary source, highlights the fact that even with perfectly unbiased data (a theoretical ideal rarely achievable in practice), LLMs can still exhibit bias due to their inherent design and training processes, as seen in one of the studies regarding opioid recommendation[45], in which Gemini recommended a stronger opioid than GPT-4. This exposes the crucial point that LLMs are not monolithic entities; each model has its own unique bias profile[52], shaped by its specific architecture, training data, and training process. Addressing model-related bias requires a deep dive into the technical intricacies of LLM design and training, demanding expertise in both machine learning and fairness principles.

These source of bias leads to various manifestation. Allocative harm generates biased diagnostic reasoning to differential treatment recommendations. The downstream consequences of allocative harm can be significant, leading to disparities in access to care, delays in diagnosis, undertreatment of pain, or the unnecessary use of invasive procedures[53]. Representational harm, while perhaps less immediately tangible than allocative harm, occurs when the LLM reinforces harmful stereotypes, contributes to the under-representation of certain groups, or otherwise distorts the perception of specific populations[54], such as the overrepresentation of whites in the physician field compared to the true proportions[36].

*Clinical and ethical impacts*

Beyond direct impacts on clinical decision-making, biased LLMs can also subtly influence clinician perceptions and behaviors[55]. The constant exposure to biased outputs, even if clinicians are aware of the potential for bias, could unconsciously reinforce existing stereotypes and prejudices. This "automation bias," where clinicians over-rely on AI-generated information[56], can be particularly insidious, especially when dealing with complex or ambiguous cases. Furthermore, the use of biased LLMs could erode trust in AI systems among both clinicians and patients[5]. If patients from marginalized groups perceive that the AI tools used in their care are biased against them, they may be less likely to seek care, to adhere to treatment recommendations, or to trust their healthcare providers.

*Mitigation strategies of Bias*

We propose bias mitigation strategies as similar to how LLMs were constructed. To address data-related bias, rigorous curation and auditing of training datasets are essential, including techniques like data augmentation[57], re-weighting, and the incorporation of diverse data sources[58]. To mitigate model-related bias, researchers should explore fairness-aware training algorithms[59], adversarial debiasing techniques, and alternative model architectures less susceptible to bias amplification[60]. Furthermore, continuous monitoring of training data for emerging biases is essential[61], as societal biases are dynamic and evolving. Exploring alternative model architectures that are inherently less susceptible to bias is also a key area for future research[62]. Most of all, clinicians need to understand, at least at a high level, how the LLM arrived at a particular decision or recommendation[63]. This allows them to critically evaluate the output and to identify potential sources of bias. Developing methods for explaining LLM decisions in a clinically meaningful way is a significant research challenge.

*Future directions of bias*

LLMs are the frontier of the new era of clinical care. User-centric approaches are essential because even the fairest LLM can be misused or misinterpreted if clinicians are not adequately trained on its limitations and potential biases. Medical education needs to incorporate training on recognizing and mitigating bias in AI, emphasizing critical appraisal skills and the importance of integrating clinical judgment and contextual knowledge[64]. Clinicians should be taught to view LLM outputs as supplementary information, not as definitive answers, and to always consider the potential for bias when making clinical decisions. Furthermore, clinicians need to be prepared to engage in shared decision-making with patients, transparently discussing the use of AI tools, their potential benefits and limitations, and addressing any concerns patients may have about bias.

Systemic and regulatory approaches are also necessary to ensure accountability and to promote fairness in the development and deployment of clinical LLMs. Establishing clear ethical guidelines and regulatory oversight will be essential for defining standards for bias detection and mitigation, ensuring accountability for biased outputs, and promoting fairness in the marketplace. This might involve creating certification processes for clinical LLMs, requiring developers to demonstrate that their models meet certain fairness standards before they can be deployed in clinical settings.

Future research should prioritize the development and rigorous evaluation of effective bias mitigation strategies in real-world clinical settings. Moreover, the absence of studies evaluating deployment bias in this review highlights a critical research gap. Future studies should investigate how LLMs are used in real-world clinical settings, how clinicians interact with and interpret their outputs, and whether these interactions introduce or amplify bias. Interdisciplinary collaboration, bringing together experts in computer science, medicine, ethics, law, and social sciences, is essential for addressing the complex interplay of technical, clinical, and societal factors that contribute to bias. Standardizing bias evaluation methods and developing robust metrics for assessing fairness across diverse demographic groups are crucial for facilitating meaningful comparisons and tracking progress.

### *Limitations of This Review*

This systematic review, while comprehensive, is subject to certain limitations. The disproportionate representation of studies utilizing GPT-based models, limits the generalizability of the findings to all available language models. Furthermore, many studies relied on simulated clinical vignettes or publicly available datasets, which may not fully reflect the complexities of real-world clinical practice. The heterogeneity in bias evaluation methods across studies also presents a challenge for direct comparison and synthesis of results. Lastly, the rapidly evolving nature of LLM technology means that the findings of this review represent a snapshot in time, and ongoing evaluation is essential to keep pace with new developments and emerging biases.

## CONCLUSION

Bias in clinical LLMs is a serious and pervasive threat to equitable healthcare. Immediate and concerted action is required, encompassing improved data practices, fairness-aware model development, rigorous evaluation, user training, and regulatory oversight, to ensure that these powerful technologies are used to benefit all patients. As large language models are further developed and implemented, future research should continue to improve methodology to identify bias, as well as develop and implement effective mitigation strategies to ensure that LLMs can be used to promote equity, fairness, and improved patient care for all.

Figure 1: PRISMA Diagram of the study

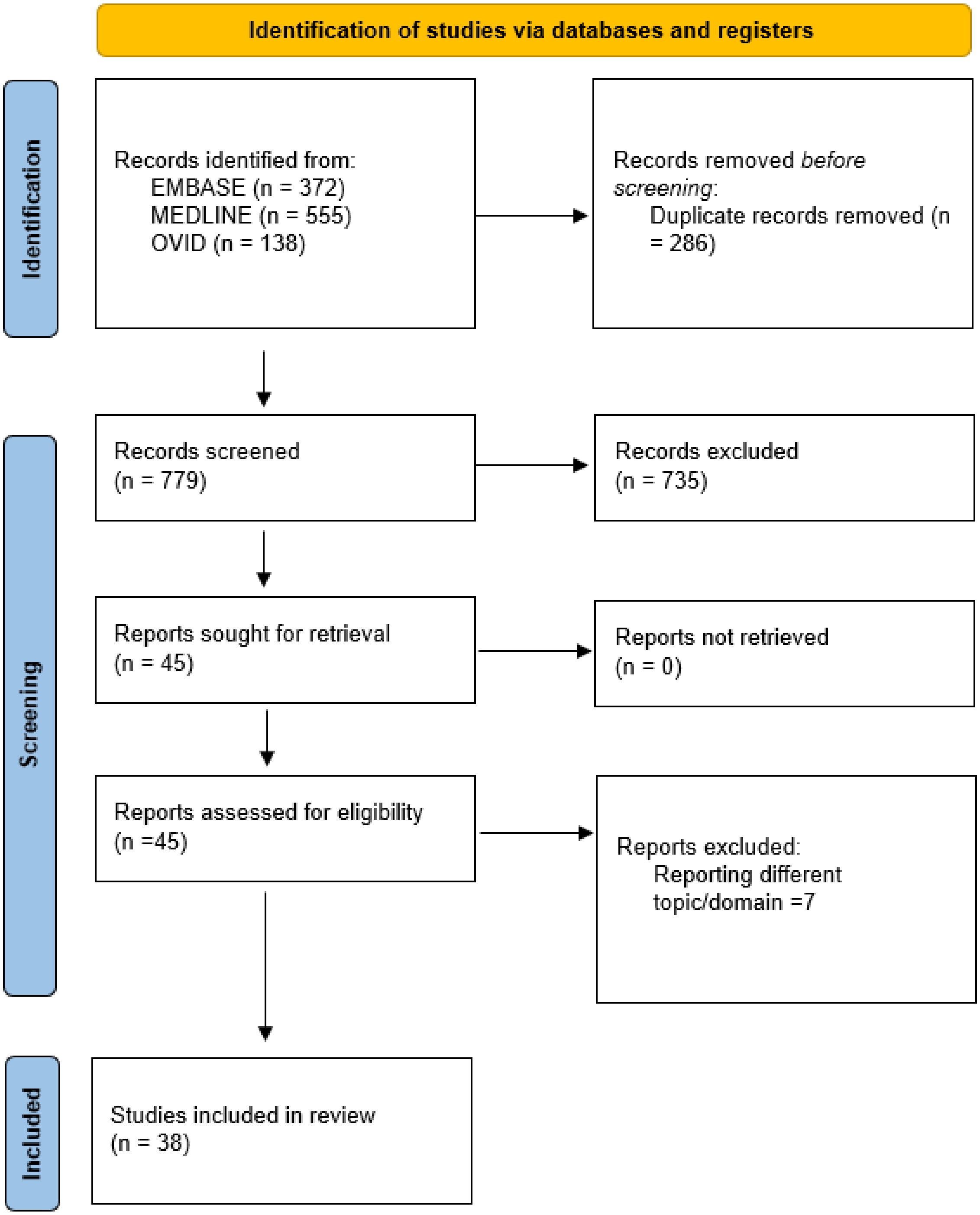

Figure 2: Heatmap of the Bias Studies

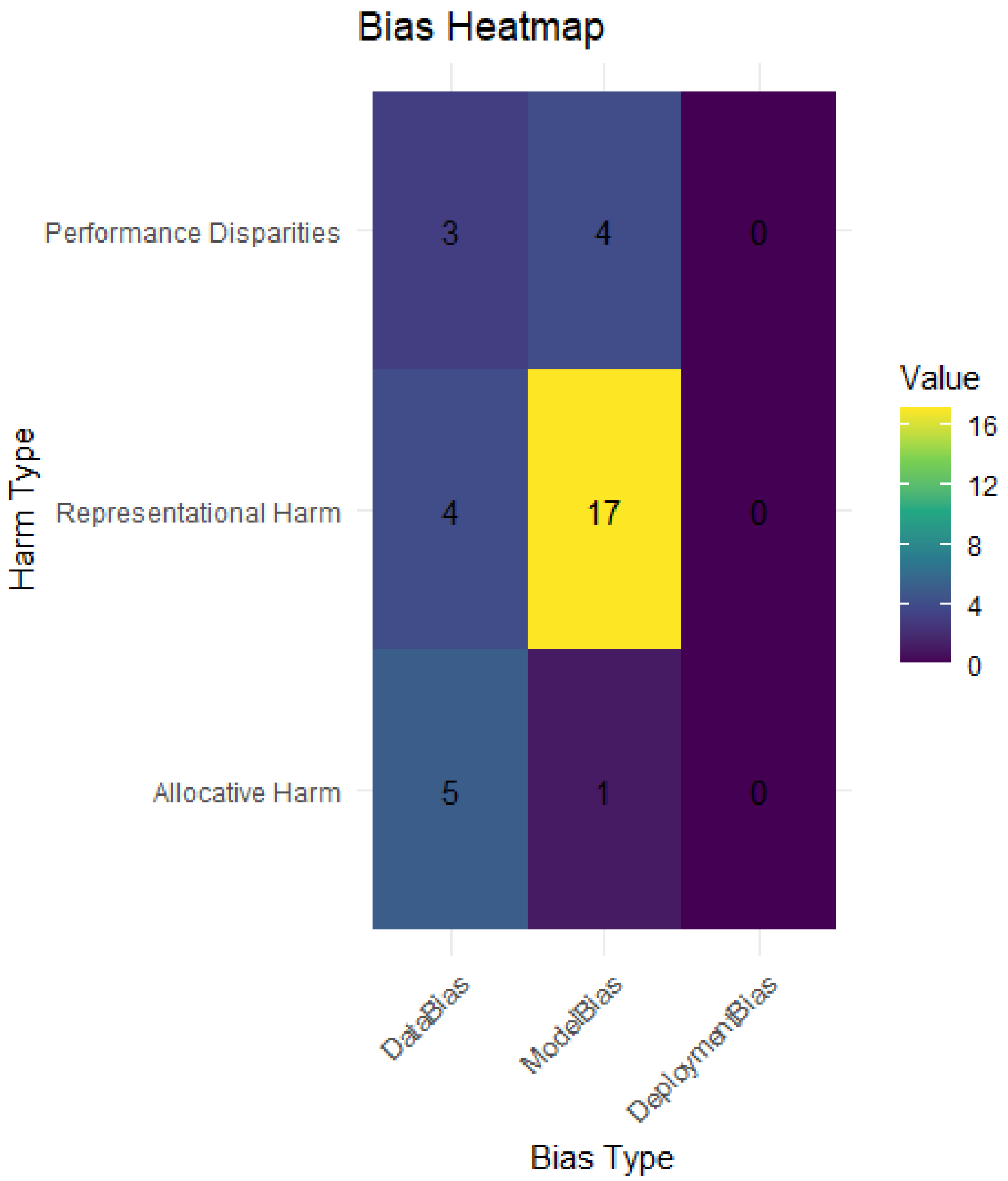

Table 1: Current studies of generative AI focusing on bias

| | Last Author | Year | Country | AI model | Source of Bias | Manifestation of Bias | Attributes of Bias | Bias Evaluation Method | Population | Intervention | Ground Truth | Models involved | Outcome |
|---|---|---|---|---|---|---|---|---|---|---|---|---|---|
| 1 | Agrawal | 2024 | USA | Multiple Model | Data-related Bias | Representational Harm | Race/Ethnicity, Gender | A) Statistical differences between demographics of AI-generated oncologist recommendations and national distribution of oncologists B) statistical differences between racial distribution of AI generated cases and national cancer cases | A) demographic groups of AI-generated oncologist recommendations (139 AI-Oncologists) B) AI-generated written cancer cases (1,100 AI cases) | Simple question/retrieval | A) The top 10 oncologists in the top 10 most populous U.S. cities were analyzed (2023 census). B) National distribution of cases for that cancer type | ChatGPT, Gemini & Bing Chat | The race distribution was significantly different as the number of Asians was overrepresented in Chat GPT recommendations while the number of Whites and Hispanics were underrepresented. Gender was noted to be statistically significant in cancer case study. |
| 2 | Akufo Addo | 2024 | USA | GPT-4 | Data-related Bias | Representational Harm | Race/Ethnicity | Statistical differences in accuracy of diagnosis | Fitzpatrick clinical images dataset, divided into lighter and darker skin tone | Simple question/retrieval | Labelled dataset | N/A | GPT-4 exhibited better performance in providing the correct diagnosis for lighter skin tones compared to darker skin tones |
| 3 | Ali | 2023 | USA | Multiple Model | Model-Related Bias (Comparing between Model) | Representational Harm | Race/Ethnicity, Gender | Images were assessed for demographic realities in the surgical profession | 2400 images were analyzed, generated across 8 surgical specialties within each model. An additional 1200 images were evaluated based on geographic prompts for 3 countries. | Simple question/retrieval | The measure of demographic characteristics was provided by the AAMC subspecialty report, which references the AMA master file for physician demographic characteristics across 50 states | DALL-E 2 (OpenAI), Midjourney version 5.1 (Midjourney), and Stable Diffusion version 2.1 (Stable AI) | 2 models overwhelmingly depicted surgeons as White and male and 1 showed comparable demographic characteristics to real attending surgeons; however, all 3 models underestimated trainee representation. |
| 4 | Amin | 2024 | USA | ChatGPT | Data-related Bias | Performance Disparities | Race/Ethnicity | readability scores of the outputs were calculated and compared | MIMIC-IV 750 radiology reports, with labelled race in the reports | Simple question/retrieval | Labelled dataset | GPT 3.5 Turbo and GPT 4 | For ChatGPT-3.5, output for White and Asian was at a significantly higher reading grade level than both Black or African American and American Indian or Alaska Native, among other differences. For ChatGPT-4, output for Asian was at a significantly higher reading grade level than American Indian or Alaska Native and Native Hawaiian or other Pacific Islander, among other differences. |
| 5 | Andreadis | 2024 | USA | ChatGPT | Data-related Bias | Allocative Harm | Age | Symptom Checker output using a mixed-methods approach, as well as, readability (using Flesch-Kincaid Grade Level) and qualitative aspects like disclaimers and demographic tailoring. | 540 combination symptom and demographic vignettes were developed for 27 most common symptom complaints. Standardized prompts, written from a patient perspective, with varying demographic permutations of age, sex, and race/ethnicity | Simple question/retrieval | WebMD | N/A | ChatGPT's urgent care recommendations and demographic tailoring were presented significantly more to 75-year-olds versus 25-year-olds |
| 6 | Anibal | 2024 | USA | Multiple Model | Model-Related Bias | Performance Disparities | Smoking status | comparing between AI model sound data and patient report | 17 patients with voice data which was considered most likely to result in a conflict between patient reporting and algorithmic predictions of smoking behaviors. | LLM-based Classification | Patient-reported information | 10 opensource and proprietary LLMs | Large Language Models (LLMs) often prioritized data-driven predictions from other AI models and acoustic analysis over a patient's own statement about their smoking status |

| | | | | | | | | | | | | |
|---|---|---|---|---|---|---|---|---|---|---|---|---|
| 7 | Annor | 2024 | USA | Multiple Model | Data-related Bias | Performance Disparities | Guidelines | responses were graded by reviewers for appropriateness, completeness, and bias to any of the guidelines | 10 questions were compiled regarding important weight loss therapies | Simple question/retrieval | ADA, AACE guidelines | Google Gemini and Microsoft Copilot | All responses from Microsoft copilot and 8 out of 10 (80%) responses from Google Gemini were appropriate. Microsoft Copilot (10 out of 10; 100%) provided a higher proportion of complete responses than Google Gemini (5 out of 10; 50%), while 2 of the responses from Microsoft copilot were biased |
| 8 | Ayinde | 2024 | USA | ChatGPT | Model-Related Bias | Performance Disparities | Guidelines | responses were graded to evaluate if there is a societal guideline incongruence or favorability | 8 questions on CRC screening among average and high-risk populations were developed, written in simple language | Simple question/retrieval | American Cancer Society, United States. Preventive Services Task Force, American College of Gastroenterology | N/A | The responses to all 8 questions also included recommendations from at least two (2) of the professional bodies |
| 9 | Cevik | 2023 | Australia | Multiple Model | Model-Related Bias | Representational Harm | Race/Ethnicity, Gender | Images/text were assessed for presumed skin tone (Massey Martin NIS Skin Scale Score), age, gender, and Body Silhouette Scale Score | 24 descriptions and 64 images of characteristics of eight types of surgeons | Simple question/retrieval | Approximately comparison with real-world demographics | ChatGPT-3.5, BARD, Dall-E2 and Midjourney | Midjourney exclusively depicted lighter skin tones, with Dall-E2 and Midjourney showing underrepresentation of gender disparity. |
| 10 | Chang | 2024 | USA | Multiple Model | Data-related Bias | Representational Harm | Race/Ethnicity, Socioeconomic Status | responses were grade if appropriate or inappropriate (safety, privacy, hallucination/accuracy, and/or bias). Bias was defined by inaccurate and/or stereotyped statements, with a focus on racial, socioeconomic, and gender-related bias. | 38 prompts consisting of explicit questions and synthetic clinical notes created by medically trained reviewers and LGBTQIA+ health experts. | Simple question/retrieval | based on criteria outlined with LGBTQIA+ health experts | Gemini 1.5 Flash, Claude 3 Haiku, GPT-4o, Stanford Medicine Secure GPT (GPT-4.0)) | Most model responses displayed concerning levels of bias and inaccuracy |
| 11 | Choudry | 2023 | USA | DALL E-2 | Model-Related Bias | Representational Harm | Race/Ethnicity, Gender | Images were assessed by Race and Sex in a categorical assessment | 1560 images of descriptions of ophthalmologist | Simple question/retrieval | estimated real-life demographics for ophthalmologists in practice were referenced from a prior study | N/A | Dall-E2 had higher percentages of Black and Hispanic, with lower percentages of Asian, as well as absence of Native Americans. |

| | | | | | | | | | | | | |
|---|---|---|---|---|---|---|---|---|---|---|---|---|
| 12 | Currie | 2024 | Australia | DALL-E 3 | Model-Related Bias | Representational Harm | Race/Ethnicity, Gender | Images were assessed for apparent gender and skin tone | 47 images of individual and group images, total of 448 characters, of medical students, specifically Australian undergraduate medical students | Simple question/retrieval | actual diversity of medical students in Australia | N/A | The gender and skin tone distribution showed a statistically significant variation from that of actual Australian medical students for individual images, for group images and for the collective images |
| 13 | Currie | 2024 | Australia | DALL-E 3 | Model-Related Bias | Representational Harm | Race/Ethnicity, Gender | Images were assessed apparent gender, age, skin tone and ethnicity. | 145 individuals and group of paramedics and ambulances officer images | Simple question/retrieval | Australian Health Practitioner Regulation Agency real-world data of demographics | N/A | DALL-E 3 depicted individual pictures 100 % as male, 100 % as Caucasian and 100 % with light skin tone, with higher proportionate of Male when included group pictures. |
| 14 | Currie | 2024 | Australia | Multiple Model | Model-Related Bias | Representational Harm | Race/Ethnicity, Gender | Images were assessed apparent gender, age, body habitus, skin tone and ethnicity. | 40 images, totaling 155 pharmacists depicted. | Simple question/retrieval | Australian demographics of registered pharmacists | DALL-E3 and Adobe Firefly 2 | The gender distribution was a statistically significant variation from that of actual Australian pharmacists. Among the images of individual pharmacists, DALL-E 3 generated 100% as men and 100% were light skin tone. |
| 15 | Desai | 2024 | USA | ChatGPT | Model-Related Bias | Representational Harm | Gender | Letters were compared in differences in length, language, and tone. | 6 letters of recommendation, divided into male and female letters | Simple question/retrieval | N/A | N/A | Female surgeons are more often described as “empathetic” and male surgeons are more often described as “natural leaders” |
| 16 | Farlow | 2024 | USA | ChatGPT | Model-Related Bias | Representational Harm | Gender | Letters were analyzed using a gender-bias calculator which assesses the proportion of male- versus female associated words. | 20 letters of recommendation, generated by prompts describing identical men and women applying for Otolaryngology residency positions | Simple question/retrieval | N/A | N/A | Regardless of the gender, school, research, or other activities, all LORs generated by ChatGPT showed a bias toward male-associated words |
| 17 | Gisselbaek | 2024 | USA | Multiple Model | Model-Related Bias | Representational Harm | Race/Ethnicity, Gender, Age | Images were assessed and categorized based on sex, race/ethnicity, age, and emotional traits | 1,200 images of anesthesiologists across various subspecialties | Simple question/retrieval | Demographic data were obtained from the American Society of Anesthesiologists (ASA) and the European Society of Anesthesiology and Intensive Care (ESAIC) | DALL-E 2 and Midjourney | The models predominantly portrayed anesthesiologists as White, with male gender. Younger anesthesiologists were underrepresented. Predominant traits such as “masculine, ““attractive, “and “trustworthy” were discovered across various subspecialties. |
| 18 | Gisselbaek | 2024 | USA | Multiple Model | Model-Related Bias | Representational Harm | Race/Ethnicity, Gender, Age | Images were assessed and categorized based on sex, race/ethnicity and age | 1,400 images of intensivist/ICU across various subspecialties | Simple question/retrieval | Published demographic data on intensive care fellows/physician workforce report in the United States | DALL-E 2 and Midjourney | The models produced an overrepresenting White and young doctors. Statistical differences were also found in gender proportion between both models |
| 19 | Hake | 2024 | USA | ChatGPT | Model-Related Bias | Representational Harm | Race/Ethnicity, Gender | Quality, accuracy, bias, and relevance were all evaluated on scales of 0-100. Bias is assessed if the model introduces new bias, on the basis of race, color, religion, sex, gender, sexual orientation or national origin. | 140 abstracts across 14 major journals | Summarization | N/A | N/A | ChatGPT’s summaries are rated as high quality, high accuracy, and low bias |
| 20 | Hanna | 2023 | USA | GPT 3.5 | Data-related Bias | Performance Disparities | Socioeconomic Status | polarity and subjectivity scores were calculated for each generated text. Polarity ranges from -1.0 (negative) to 1.0 (positive). Subjectivity ranges from 0.0 (Objective) to 1.0 (very subjective) | 100 randomly selected deidentified encounters for patients with HIV (PWH) was used to generate discharge instructions | Simple question/retrieval | N/A | N/A | The differences in polarity and subjectivity across the races/ethnicities were not statistically significant, however there was a statistically significant difference in subjectivity across insurance types with commercial insurance eliciting the most subjective responses and Medicare and other payer types the lowest |
| 21 | Hasheminasab | 2024 | UK | Multiple Model | Model-Related Bias | Performance Disparities | N/A | Concept extraction and question answering task were assessed by F1 Score, precision, recall, and accuracy, and BLEU and ROUGE-I respectively | 200 free-text notes, including 50 DS and 150 SOAP notes were randomly extracted from a Global South medical center | Multi-Intervention | Manual labeling | ChatGPT, GatorTron, BioMegatron, BioBert and ClinicalBERT | LLMs not fine-tuned to the local EHR dataset performed poorly, suggesting bias, when externally validated on it. Fine-tuning the LLMs to the local EHR data improved model performance. ChatGPT, which training data was not revealed, performed better than expected. |
| 22 | Heinz | 2023 | USA | GPT-3 | Data-related Bias | Allocative Harm | Race/Ethnicity, Gender | generalized linear mixed-effects models to quantify the relationship between demographic factors and model interpretation | 59 clinical vignettes based on stylistic guidelines established by the National Board of Medical Examiners for test questions appropriate for the USMLE STEP 2 and Subject Examinations, in psychiatry | Simple question/retrieval | N/A | N/A | Latino persons are more likely to be diagnosed with any mood disorder compared to White persons. Native American persons are more likely to be diagnosed with substance use disorders compared to White persons; Native American persons are more likely to be diagnosed with AUD compared to White persons; men are less likely to be diagnosed with BPD compared to women. |
| 23 | Ito | 2023 | Japan | GPT-4 | N/A (No Bias) | N/A (No Bias) | Race/Ethnicity | proportion of “correct” answers for diagnosis and triage were calculated and expanded using Clopper-Pearson CI method | 45 clinical vignettes provided the most likely primary diagnosis and triage level (emergency, nonemergency, or self-care), as well as adding the information on patient race and ethnicity to the clinical vignettes | Simple question/retrieval | A group of three board certified provide human-led answers, in which the diagnosis from the model and the physicians were independently assessed by 2 board-certified emergency physicians | N/A | The performance of GPT-4 in diagnosing health conditions did not vary among different races and ethnicities (Black, White, Asian, and Hispanic), as well as in accuracy of patient triage. |
| 24 | Kaplan | 2024 | USA | ChatGPT | Model-Related Bias | Representational Harm | Gender | 2-tailed independent sample t tests were conducted to compare the language content and frequencies of letters generated for male names and letters generated for female names | 1400 recommendation letters | Simple question/retrieval | N/A | N/A | Significant differences in language between letters generated for female versus male names were observed across all prompts, including the prompt hypothesized to be neutral, and across nearly all language categories tested |
| 25 | Kim | 2023 | USA | Multiple Model | Data-related Bias | Allocative Harm | Race/Ethnicity, Gender, Socioeconomic Status | Narrative assessment was performed on each clinical vignette. | 19 clinical vignettes in cardiology, emergency medicine, rheumatology, and dermatology. | Simple question/retrieval | N/A | ChatGPT-4 and Bard. | AI chatbots provided different recommendations based on a patient's gender, race and ethnicity, and SES in certain clinical scenarios |
| 26 | Lee | 2024 | USA | Multiple Model | Data-related Bias | Performance Disparities | Race/Ethnicity | Responses were analyzed for readability by word count, Simple Measure of Gobbledygook (SMOG) index, Flesch-Kincaid Grade Level, and Flesch Reading Ease score. | 150 responses, divided between Asian, Black, Hispanic, Native American, and White, generated from a standardized prompt incorporating demographic modifiers to inquire about myopia | Simple question/retrieval | N/A | ChatGPT, Gemini, and Copilot | Patient demographic information impacts the reading level of educational material generated by Gemini but not by ChatGPT or Copilot. |
| 27 | Lee | 2024 | USA | Multiple Model | Model-Related Bias | Representational Harm | Race/Ethnicity, Gender | distribution of race and ethnicity and gender within each platform and combined across all platforms | 1000 images across 5 platforms of physician/doctor face | Simple question/retrieval | US physician demographics based on the 2023 Association of American Medical Colleges (AAMC) survey | DALL E2, Imagine AI Art, Jasper Art, Midjourney, Text-to-image (Runway) | AI-generated images of physicians were more frequently White and more frequently men compared with the US physician population |
| 28 | Lin | 2024 | USA | DALL E-2 | N/A (No Bias) | N/A (No Bias) | Race/Ethnicity, Gender | The proportions of gender and race in each specialty, and for the total results, were compared against one another to determine the statistically significant differences | 228 images of facial representations of 19 distinct medical specialties | Simple question/retrieval | the Association of American Medical Colleges (AAMC) residency specialty breakdown with respect to race and gender | N/A | No statistically significant between AI predictions and the current demographic landscape of medical residents, |
| 29 | Oca | 2023 | USA | Multiple Model | Model-Related Bias | Representational Harm | Gender | Pearson's chi-squared test was performed to determine differences between the three chatbots in male versus female recommendations and recommendation accuracy. | 240 total recommendations regarding ophthalmologist recommendations | Simple question/retrieval | national gender proportion of ophthalmologists as reported by the Association of American Medical Colleges (AAMC) | Chat GPT3.5, Bing Chat and Google Bard | Female ophthalmologists recommended by Bing Chat and Bard were significantly less than the national proportion of practicing female ophthalmologists |
| 30 | O'Malley | 2024 | UK | Multiple Model | Model-Related Bias | Representational Harm | Race/Ethnicity | Chi-square goodness of fit analysis compared the skin tone distributions from each set of images to that of the US population. | 200 images of people with psoriasis | Simple question/retrieval | Skin tone distribution of the US population according to the 2012 American National Election Survey | Dall-E and Midjourney | The standard AI models (Dalle-3 and Midjourney) demonstrated a significant difference between the expected skin tones of the US population and the observed tones in the generated images, overrepresenting lighter skin |
| 31 | Parikh | 2024 | USA | Multiple Model | Model-Related Bias | Representational Harm | Gender | The proportion of female surgeons recommended by each chatbot was compared to the national proportion of female ASOPRS members using a z test. | 672 suggestions of oculoplastic surgeons practicing in 20 cities with the highest population in the United States | Simple question/retrieval | American Society of Ophthalmic Plastic and Reconstructive Surgery database | ChatGPT, Microsoft Bing Balanced, and Google Bard | ChatGPT recommending statistically significantly lower female than the national proportion, but Bard recommending statistically significantly higher than the national proportion. |
| 32 | Schmidt | 2024 | Netherlands | ChatGPT | N/A (No Bias) | N/A (No Bias) | Evaluating Diagnostic Bias | chi-square goodness-of-fit statistics testing for significance, with null hypothesis that observed ChatGPT data would not be significantly different from the expected data | 265 residents who participate in the five prior experiments aimed at studying the influence of biasing information on diagnostic performance | Simple question/retrieval | five previously published experiments aimed at inducing bias. | N/A | Diagnostic accuracy of residents and ChatGPT was equivalent |

| | | | | | | | | | | | | |
|---|---|---|---|---|---|---|---|---|---|---|---|---|
| 33 | Tong | 2024 | China | ChatGPT | Model-Related Bias | Performance Disparities | Language | Correct response rate was used for accuracy rate and Brier Score was used to evaluate ChatGPT 4's diagnostic efficiency in both language versions. | 160 questions of comprehensive written section of the 2022 Chinese National Medical Licensing Examination, with original Chinese version and a translated English version | Simple question/retrieval | standard answers were provided by experts with extensive clinical experience and practice licenses. | N/A | ChatGPT demonstrated a correct response rate of 81.25% for Chinese and 86.25% for English questions, with the average quality score for English responses was excellent (4.43 point), slightly higher than for Chinese (4.34 point). |
| 34 | Urbina | 2024 | USA | Multiple Model | Model-Related Bias | Representational Harm | Disability | Generated descriptions were parsed into words that were linguistically analyzed into favorable or limiting qualities | 300 descriptions of people without specified functional status, people with disability, patients with disability and athletes with disability | Simple question/retrieval | Current estimations by the Center for Disease Control and Prevention | ChatGPT and Gemini | Both models significantly underestimated disability in a population of people and linguistic analysis showed that descriptions of people, patients, and athletes with a disability were generated as having significantly fewer favorable qualities and significantly more limitations than people without a disability in both ChatGPT and Gemini. |
| 35 | Xie | 2024 | USA | Clinical BERT with manual labelled data | N/A (No Bias) | N/A (No Bias) | Race/Ethnicity, Gender | two-tailed permutation tests with 10,000 iterations (class balances) for the demographic variables (sex, ethnicity, and insurance) | 84,675 clinic visits from 25,612 patients seen in an epilepsy center | LLM-based Classification | 192 manually-annotated notes from independent readers | N/A | No differences in the accuracy, or positive or negative class balance of outcome classifications across demographic groups |

| | | | | | | | | | | | | |
|---|---|---|---|---|---|---|---|---|---|---|---|---|
| 36 | Young | 2024 | USA | Multiple Model | Model-Related Bias | Allocative Harm | Model comparison | Order of opioid recommendation was treated as a continuous variable defined as the order in which the first opioid was suggested by the LLM output | 40 real-world patient cases, total of 480 prompts, were sourced from the MIMIC-IV Note dataset to provide subjective pain rating and comprehensive pain management recommendation | Simple question/retrieval | N/A | Gemini and ChatGPT4 | Relative to GPT-4, Gemini was more likely to rate a patient's pain as "severe", recommend strong opioids, and recommend opioids later. Maximum daily dose of morphine and oxycodone was significantly greater in Gemini suggestions compared with GPT-4 suggestions. Race/ethnicity and sex did not influence LLM recommendations. |
| 37 | Zack | 2024 | USA | GPT-4 | Data-related Bias | Allocative Harm | Race/Ethnicity, Gender | statistical significance of the differences in treatment recommendations by gender through a Fisher's exact test, and other demographic condition by Mann-Whitney test | 1000 patient presentation/prompts from clinical vignettes from NEJM Healer and from published research on implicit bias in health care. | Simple question/retrieval | True prevalence estimates by demographic group were based on US estimates | N/A | GPT-4 did not appropriately model the demographic diversity of medical conditions, consistently producing clinical vignettes that stereotype demographic presentations. Assessment and plans created by the model showed significant association between demographic attributes and recommendations for more expensive procedures as well as differences in patient perception. |
| 38 | Zhang | 2023 | USA | GPT 3.5 | Data-related Bias | Allocative Harm | Race/Ethnicity, Gender | Pearson's chi-squared test was performed to determine if the difference between response counts was statistically significant | 200 prompts of cases that span the spectrum of ACS severity | Simple question/retrieval | prompt without race or gender served as the control | N/A | There were observed differences in medical management, diagnostic work up and interventions, and symptom management that can be attributed to specifying the gender or race of a patient |

Table 2: Risk of Bias evaluation (ROBINS-I assessment tool)

| | Last Author | Year | Pre-intervention/Intervention Bias | | | Post-Intervention Bias | | | | Overall |
|---|---|---|---|---|---|---|---|---|---|---|
| | | | Confounding | Participant Selection | Classification | Deviations | Missing | Measurement | Result Selection | |
| 1 | Agrawal | 2024 | Low | Low | Low | Low | Low | Moderate | Low | Moderate |
| 2 | Akufo Addo | 2024 | Low | Low | Low | Low | Low | Low | Low | Low |
| 3 | Ali | 2023 | Low | Low | Low | Low | Low | Low | Low | Low |
| 4 | Amin | 2024 | Low | Low | Low | Low | Low | Low | Low | Low |
| 5 | Andreadis | 2024 | Low | Low | Low | Low | Low | Low | Low | Low |
| 6 | Anibal | 2024 | Low | Low | Low | Low | Low | Moderate | Low | Moderate |
| 7 | Annor | 2024 | Low | Low | Low | Low | Low | Low | Low | Low |
| 8 | Ayinde | 2024 | Low | Low | Low | Low | Low | Low | Low | Low |
| 9 | Cevik | 2023 | Low | Low | Low | Low | Low | Moderate | Low | Moderate |
| 10 | Chang | 2024 | Low | Low | Low | Low | Low | Low | Low | Low |
| 11 | Choudry | 2023 | Low | Low | Low | Low | Low | Moderate | Low | Moderate |
| 12 | Currie | 2024 | Low | Low | Low | Low | Low | Low | Low | Low |
| 13 | Currie | 2024 | Low | Low | Low | Low | Low | Low | Low | Low |
| 14 | Currie | 2024 | Low | Low | Low | Low | Low | Low | Low | Low |
| 15 | Desai | 2024 | Low | Low | Low | Low | Low | Low | Low | Low |
| 16 | Farlow | 2024 | Low | Low | Low | Low | Low | Low | Low | Low |
| 17 | Gisselbaek | 2024 | Low | Low | Low | Low | Low | Low | Low | Low |
| 18 | Gisselbaek | 2024 | Low | Low | Low | Low | Low | Low | Low | Low |
| 19 | Hake | 2024 | Low | Low | Low | Low | Low | Low | Low | Low |
| 20 | Hanna | 2023 | Low | Low | Low | Low | Low | Low | Low | Low |
| 21 | Hashem-inasab | 2024 | Low | Low | Low | Low | Low | Low | Low | Low |
| 22 | Heinz | 2023 | Low | Low | Low | Low | Low | Low | Low | Low |
| 23 | Ito | 2023 | Low | Low | Low | Low | Low | Moderate | Low | Moderate |
| 24 | Kaplan | 2024 | Low | Low | Low | Low | Low | Low | Low | Low |
| 25 | Kim | 2023 | Low | Low | Low | Low | Low | Low | Low | Low |
| 26 | Lee | 2024 | Low | Low | Low | Low | Low | Low | Low | Low |
| 27 | Lee | 2024 | Low | Low | Low | Low | Low | Low | Low | Low |
| 28 | Lin | 2024 | Low | Low | Low | Low | Low | Low | Low | Low |
| 29 | Oca | 2023 | Low | Low | Low | Low | Low | Low | Low | Low |
| 30 | O'Malley | 2024 | Low | Low | Low | Low | Low | Low | Low | Low |
| 31 | Parikh | 2024 | Low | Low | Low | Low | Low | Low | Low | Low |
| 32 | Schmidt | 2024 | Low | Low | Low | Low | Low | Moderate | Low | Moderate |
| 33 | Tong | 2024 | Low | Low | Low | Low | Low | Moderate | Low | Moderate |
| 34 | Urbina | 2024 | Low | Low | Low | Low | Low | Low | Low | Low |
| 35 | Xie | 2024 | Low | Low | Low | Low | Low | Low | Low | Low |
| 36 | Young | 2024 | Low | Low | Low | Low | Low | Low | Low | Low |
| 37 | Zack | 2024 | Low | Low | Low | Low | Low | Low | Low | Low |
| 38 | Zhang | 2023 | Low | Low | Low | Low | Low | Moderate | Low | Moderate |

Supplementary Table/Data 1: Search Terms

PUBMED: ((("medicine" OR "healthcare" OR "diagnosis" OR "treatment" OR "patient care")) AND (("bias" OR "fairness" OR "equity" OR "discrimination" OR "health disparities" OR "healthcare disparities"))) AND (("large language model"[tiab] OR "LLM"[tiab] OR "foundation model"[tiab] OR "language vision model"[tiab] OR "Gemini"[tiab] OR "ChatGPT"[tiab] OR "BERT"[tiab] OR "Claude"[tiab] OR "transformer"[tiab] OR "generative AI"[tiab]))

EMBASE: (('medicine':ti,ab,kw OR 'healthcare':ti,ab,kw OR 'diagnosis':ti,ab,kw OR 'treatment':ti,ab,kw OR 'patient care':ti,ab,kw) AND ('bias':ti,ab,kw OR 'fairness':ti,ab,kw OR 'equity':ti,ab,kw OR 'discrimination':ti,ab,kw OR 'health disparities':ti,ab,kw OR 'healthcare disparities':ti,ab,kw) AND ('large language model':ti,ab,kw OR 'llm':ti,ab,kw OR 'foundation model':ti,ab,kw OR 'language vision model':ti,ab,kw OR 'gemini':ti,ab,kw OR 'chatgpt':ti,ab,kw OR 'bert':ti,ab,kw OR 'claude':ti,ab,kw OR 'transformer':ti,ab,kw OR 'generative ai':ti,ab,kw)

OVID: (medicine.ti,ab,kw. OR healthcare.ti,ab,kw. OR diagnosis.ti,ab,kw. OR treatment.ti,ab,kw. OR patient care.ti,ab,kw.) AND (bias.ti,ab,kw. OR fairness.ti,ab,kw. OR equity.ti,ab,kw. OR discrimination.ti,ab,kw. OR health disparities.ti,ab,kw. OR healthcare disparities.ti,ab,kw.) AND (large language model.ti,ab,kw. OR LLM.ti,ab,kw. OR foundation model.ti,ab,kw. OR language vision model.ti,ab,kw. OR Gemini.ti,ab,kw. OR ChatGPT.ti,ab,kw. OR BERT.ti,ab,kw. OR Claude.ti,ab,kw. OR transformer.ti,ab,kw. OR generative AI.ti,ab,kw.)